\documentclass[10pt,twocolumn,letterpaper]{article}

\usepackage{cvpr}              % To produce the CAMERA-READY version
\usepackage{graphicx}
\usepackage{amsmath}
\usepackage{amssymb}
\usepackage{booktabs}
\usepackage{makecell}
\usepackage{times}
\usepackage{epsfig}
\usepackage{graphicx}
\usepackage{amsmath}
\usepackage{amssymb}
\usepackage[accsupp]{axessibility}
\usepackage[numbers, sort]{natbib}
\usepackage{helvet}   %Required
\usepackage{courier}  %Required
\usepackage{url}      %Required
\usepackage{xcolor}
\usepackage{tabularx}
\usepackage{booktabs}
\usepackage{multirow}
\usepackage{subcaption}
\usepackage{comment}
\usepackage{algorithmic}
\usepackage[export]{adjustbox}
\usepackage{color}
\usepackage{pifont}
\usepackage{enumitem}

\usepackage[pagebackref,breaklinks,colorlinks]{hyperref}

\usepackage[capitalize]{cleveref}
\crefname{section}{Sec.}{Secs.}
\Crefname{section}{Section}{Sections}
\Crefname{table}{Table}{Tables}
\crefname{table}{Tab.}{Tabs.}

\def\cvprPaperID{10} % *** Enter the CVPR Paper ID here
\def\confName{CVPR}
\def\confYear{2023}

\makeatletter

\newcommand*\titleheader[1]{\gdef\@titleheader{#1}}
\AtBeginDocument{%
  \let\st@red@title\@title
  \def\@title{%
    \vspace{-1.2cm}
    \bgroup\normalfont\footnotesize\centering\textcolor{gray}{\@titleheader}\par\egroup
    \vskip1.5em\st@red@title}
}

\makeatother

\title{Multiface: A Dataset for Neural Face Rendering}
\titleheader{Published on CVPR 2023 Workshop \\ 3DMV: Learning 3D with Multi-View Supervision}

\author{
Cheng-hsin Wuu$^*$,
Ningyuan Zheng\thanks{Equal contribution.} ,
Scott Ardisson,
Rohan Bali,
Danielle Belko,
Eric Brockmeyer, \\
Lucas Evans,
Timothy Godisart,
Hyowon Ha,
Xuhua Huang,
Alexander Hypes,
Taylor Koska,
Steven Krenn, \\
Stephen Lombardi,
Xiaomin Luo,
Kevyn McPhail,
Laura Millerschoen,
Michal Perdoch,
Mark Pitts, \\
Alexander Richard,
Jason Saragih,
Junko Saragih,
Takaaki Shiratori,
Tomas Simon,
Matt Stewart, \\
Autumn Trimble,
Xinshuo Weng,
David Whitewolf,
Chenglei Wu,
Shoou-I Yu,
Yaser Sheikh \vspace{0.5cm} \\
Meta Reality Labs Research \\
{\tt\small \{ecwuu, zhengningyuan\}@meta.com}}

\begin{document}

\maketitle

%%%%%%%%% ABSTRACT
\begin{abstract}
% In this work, we present a new multi-view, high-resolution human face dataset of unprecedented detail and quality.
% We collect a large set of more than 100 facial expressions each from 13 identities with dozens of camera views at high resolution.

% To keep the consistency in lighting condition and distribution of facial expression, we place hundreds of directional LED point lights directed at the face to produce uniform illumination and have each subject make facial expressions from a predefined set of over a hundred expressions.

Photorealistic avatars of human faces have come a long way in recent years, yet research along this area is limited by a lack of publicly available, high-quality datasets covering both, dense multi-view camera captures, and rich facial expressions of the captured subjects. In this work, we present Multiface, a new multi-view, high-resolution human face dataset collected from 13 identities at Reality Labs Research for neural face rendering. We introduce Mugsy, a large scale multi-camera apparatus to capture high-resolution synchronized videos of a facial performance. The goal of Multiface is to close the gap in accessibility to high quality data in the academic community and to enable research in VR telepresence. Along with the release of the dataset, we conduct ablation studies on the influence of different model architectures toward the model's interpolation capacity of novel viewpoint and expressions. With a conditional VAE model \cite{10.1145/3197517.3201401} serving as our baseline, we found that adding spatial bias, texture warp field, and residual connections improves performance on novel view synthesis. Our code and data is available at: \url{https://github.com/facebookresearch/multiface}. 
\end{abstract}

%%%%%%%%% BODY TEXT
\section{Introduction}
\label{sec:intro}

\begin{figure*}[t]
  \centering
  \includegraphics[width=0.8\linewidth]{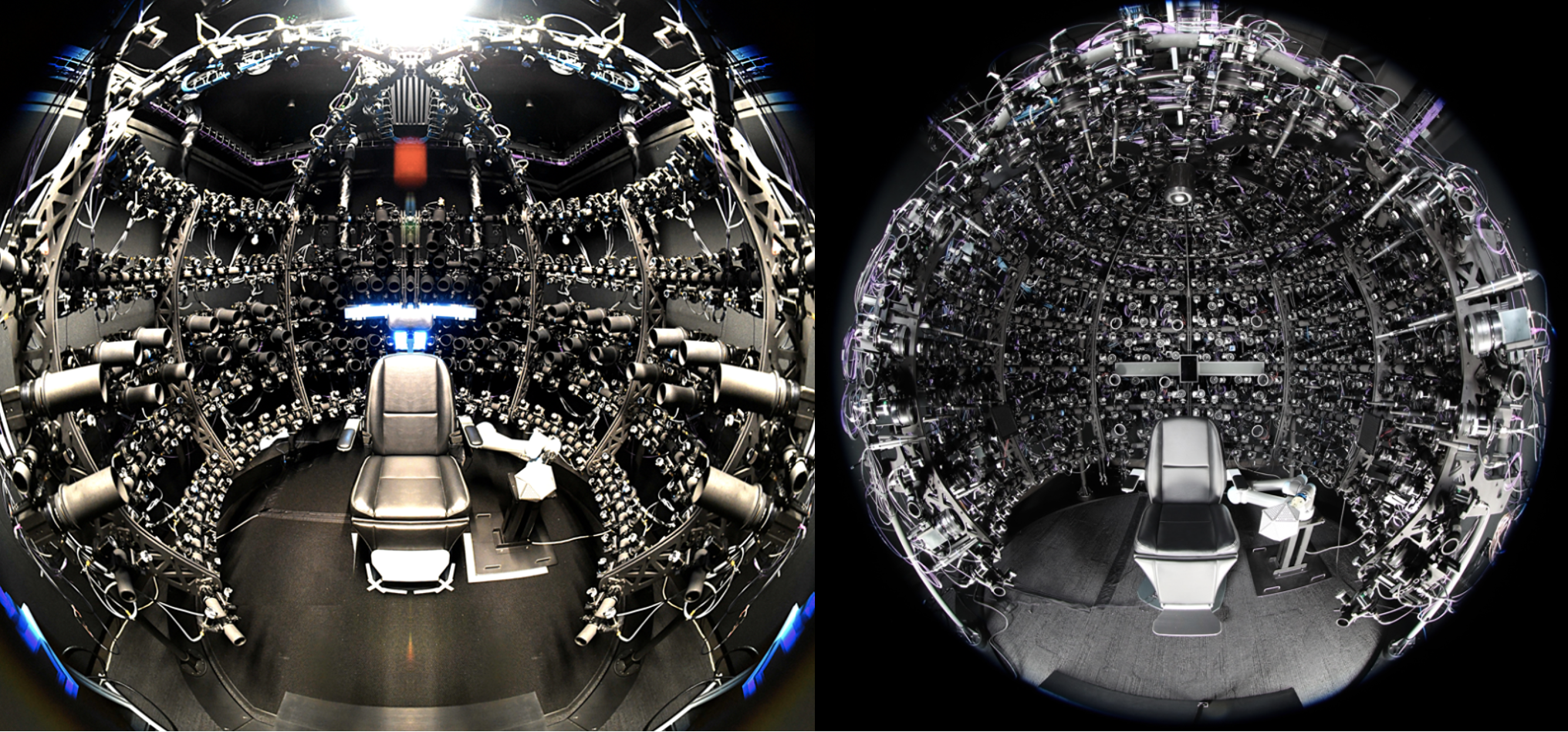}
  \caption{Mugsy v1 (left) and Mugsy v2 (right).}
	\label{fig: mugsy_dome}
\end{figure*}

Photo-realistic human face rendering and reconstruction is essential to real-time telepresence technology that drives modern Virtual Reality applications. Since humans are social animals that have evolved to express and read emotions from subtle changes in facial expressions, tiny artifacts give rise to the uncanny valley that could hurt user experience. Nowadays, many modern 3D telepresence methods leverage deep learning models and neural rendering for high-fidelity reconstruction, and to tackle difficult problems such as novel view synthesis and view-dependent effects modeling ~\cite{nerf, neuralvolumes, dnerf, nerfie}. These approaches are usually data-hungry, and the design of a capture system and data collection pipeline directly determines the performance of those models.
Pushing the boundaries in such photo-realistic human face models therefore requires a large dataset of high-resolution, multi-view facial images spanning a wide variety of expressions.
We introduce such a dataset, collected by a high-end multi-view capturing system (\textit{Mugsy}) that we
built at Meta Reality Labs Research in Pittsburgh. Compared to existing face modeling datasets such as HUMBI ~\cite{yu2020humbi} and FaceWarehouse~\cite{6654137}, our Codec-Avatar dataset contains facial data of unprecedented quality, variation in facial expressions, and number of camera views.
We capture 13 subjects with a great variety of high-fidelity facial expressions along with the geometry mesh tracked over the capturing time. For each subject, we have over a hundred facial expressions captured by multiple machine vision cameras synchronously at 4096$\times$2668 resolution (about 11 megapixels).
The capture system is illustrated in Figure~\ref{fig: mugsy_dome}.

We release the images of all 13 captured participants as well as tracked meshes and unwrapped textures, along with camera calibrations and audio data.
Moreover, we provide code to train a Codec Avatar from scratch as well as pre-trained Codec Avatars for all identities.

The creation of data-driven avatars is challenging along three major axes: (1) novel view synthesis, as we could not have cameras placed everywhere (2) novel expression synthesis, as we could not ask the participants to enact all possible facial expressions during the capture, and (3) relighting, as it is impossible to capture every possible lighting configuration.
In this report, we focus on the first two axes, while relighting is beyond the scope of this work.
We use a conditional VAE model~\cite{10.1145/3197517.3201401} as a baseline, and evaluate the model's reconstruction quality with respect to different network architectures, which includes spatial biases, a texture warp field, and residual blocks.
Empirically, we found that the baseline model benefits from these architectural modifications in interpolating novel views.

After addressing related work, we provide details on our capture system \textit{Mugsy} in Section~\ref{sec:dataset_creation} and the capture script and process used to collect the dataset. In Section~\ref{sec:technical}, we describe the model architectures and training pipeline for building the photo-realistic Codec Avatar. In Section~\ref{sec:results}, we present an ablation study of how different model architectures respond to synthesizing viewpoints and expressions.
%while we do not observe the same improvement of model's performance on generalizing to unseen expressions.

%------------------------------------------------------------------------
\section{Related Works}
\label{sec:related_works}

We briefly review two prior efforts in human face data collection, HUMBI~\cite{yu2020humbi} and FaceWarehouse~\cite{6654137}, and compare these datasets with our dataset, see Table~\ref{table: dataset comparision}.

\paragraph{HUMBI~\cite{yu2020humbi}} is a large-scale multi-view dataset designed to facilitate high resolution pose- and view-specific appearance of human body expressions (gaze, face, hand, body, and garment). The database uses a dense camera array composed of 107 synchronized cameras to capture 772 distinctive subjects doing diverse activities. The presence of HUMBI shows a new opportunity to build a versatile model that generates data-driven photo-realistic rendering for full body avatars. While HUMBI focuses on capturing multi-view images for the entire body, it does not provide high enough resolution for the images.
This restriction poses a challenge on reconstructing the subtle changes of human facial expression, which we aim to overcome with our dataset.

\begin{table}[tb]
  \centering
  \footnotesize
  \caption{Comparison of multi-view datasets.}
  \label{table: dataset comparision}
  \begin{tabularx}{0.45\textwidth}{Xrrr}
    \toprule
    Dataset & Camera Resolution & $\#$ Expressions & $\#$ View \\
    \midrule
    FaceWarehouse  & 640x480 & 20 & 1\\
    HUMBI & 1920x1080 & 20& 32\\
    \midrule
    Mugsy v1	& 2048x1334 & 65 & 40 \\
    Mugsy v2	 & 2048x1334	&	118 & 150   \\
    \bottomrule
  \end{tabularx}
\end{table}

\paragraph{FaceWarehouse~\cite{6654137}} is a database of 3D facial expressions for visual computing applications. The database uses Kinect, an off-the-shelf RGBD camera, to capture and estimate facial geometry and texture of 150 subjects, each with 20 expressions. Compared with previous 3D facial databases, FaceWarehouse has richer collections of expressions for each person that enables depiction of most human facial actions. This dataset has potential on applications such as facial image manipulation, face component transfer, real-time performance-based facial image animation, and facial animation retargeting from video to image. However, the data in FaceWarehouse does not contain detailed facial geometries such as wrinkles due to the low precision in depth information provided from the capture apparatus. This insufficiency makes applications such as high-fidelity 3D facial reconstruction very challenging.

Our dataset, in contrast, provides the richness in facial expressions together with high-resolution images that enables us to model nuanced yet important subtleties in human faces up to the level of skin pores.

\paragraph{Existing works on Codec Avatars.}
In the past, a line of works emerged from our lab that is built on the data released in this dataset.
Deep appearance models~\cite{10.1145/3197517.3201401} are the first successful approach to building photorealistic, high quality avatars but suffer from limitations in their expressiveness and controllability.
To overcome these issues, in~\cite{chu2020expressive}, the original deep appearance model has been replaced by a fully convolutional architecture that aims at increased modularity among different facial regions and thereby achieves higher expressiveness.
Focusing on the importance of eye contact in human communication, \cite{schwartz2020eyes} extends Codec Avatars with an explicit eye model.
A limitation of all above approaches is their reliance on traditional mesh-based rendering, which falls short in its ability to accurately render thin structures, translucency, and biological motion.
With the uprise of neural rendering, this reliance on mesh-based rendering has been largely overcome, and avatars have been shown to expose outstanding details and realistic modeling of difficult regions such as hair~\cite{neuralvolumes, wang2021learning, lombardi2021mixture, nam2019strand, sun2021human}.
Audio-visual data has been leveraged in~\cite{richard2021meshtalk, richard2021audio}, both in a mesh-based setup as well as in fully-textured avatar animation from audio and gaze.
Finally, Pixel Codec Avatars~\cite{ma2021pixel} provide a lightweight model that can render photorealistic avatars on commodity hardware such as a Quest 2.

%------------------------------------------------------------------------
\section{Dataset Creation}
\label{sec:dataset_creation}
In this section, we detail how the dataset was created, starting with an overview of the dataset characteristics, followed by a capture system description, a description of the capture script participants went through, and the tracking pipeline used to process the captured data.

% \alex{
% Readers are lazy. The first thing they want to know are the high-level dataset statistics so they can determine if this is the kind of data they are looking for:
% \begin{itemize}
%     \item Dataset size: 13 subjects, each captured in a multi-view capture stage performing various facial expressions, average of X frames per subject captured at 30 (90?) fps (do we release the 90 fps data or the 30 fps data?), each frame from 40 (v1) to 160 (v2) different camera views
%     \item provided assets: raw images, tracked meshes, unwrapped textures, metadata such as camera calibration and headpose, audio (?)
%     \item code to build a Codec Avatar as described in~\cite{10.1145/3197517.3201401}
% \end{itemize}
% }

\subsection{Dataset Overview}

Our dataset consists of high quality recordings of the faces of 13 identities, each captured in a multi-view capture stage performing various facial expressions. An average of 12,200 (capture version 1) to 23,000 (capture version 2) frames per subject were captured at 30 fps. Each frame has roughly 40 (v1) or 150 (v2) different camera views under uniform illumination.
%, yielding a total dataset size of 400TB\footnote{Note that due to the size of the dataset, we also provide precomputed textures and meshes to reduce the data load for the users of our dataset.}.

We provide the captured images from each camera view at a resolution of 2048 $\times$ 1334 pixels, tracked meshes including headposes, unwrapped textures at 1024 $\times$ 1024 pixels, metadata including intrinsic and extrinsic camera calibrations, and audio.
Additionally, we release code to download the dataset and build a Codec Avatar using a deep appearance model~\cite{10.1145/3197517.3201401}.
All required code and dataset documentation will be publicly available online\footnote{\url{https://github.com/facebookresearch/multiface}}.

\subsection{Capture Studio}
In order to capture synchronized multi-view videos of a facial performance, we built a multi-video-camera capture dome called \textit{Mugsy} (short from \textit{Mugshooter}), see Figure~\ref{fig: mugsy_dome}.

\begin{figure*}
    \centering
    \includegraphics[width=0.8\linewidth]{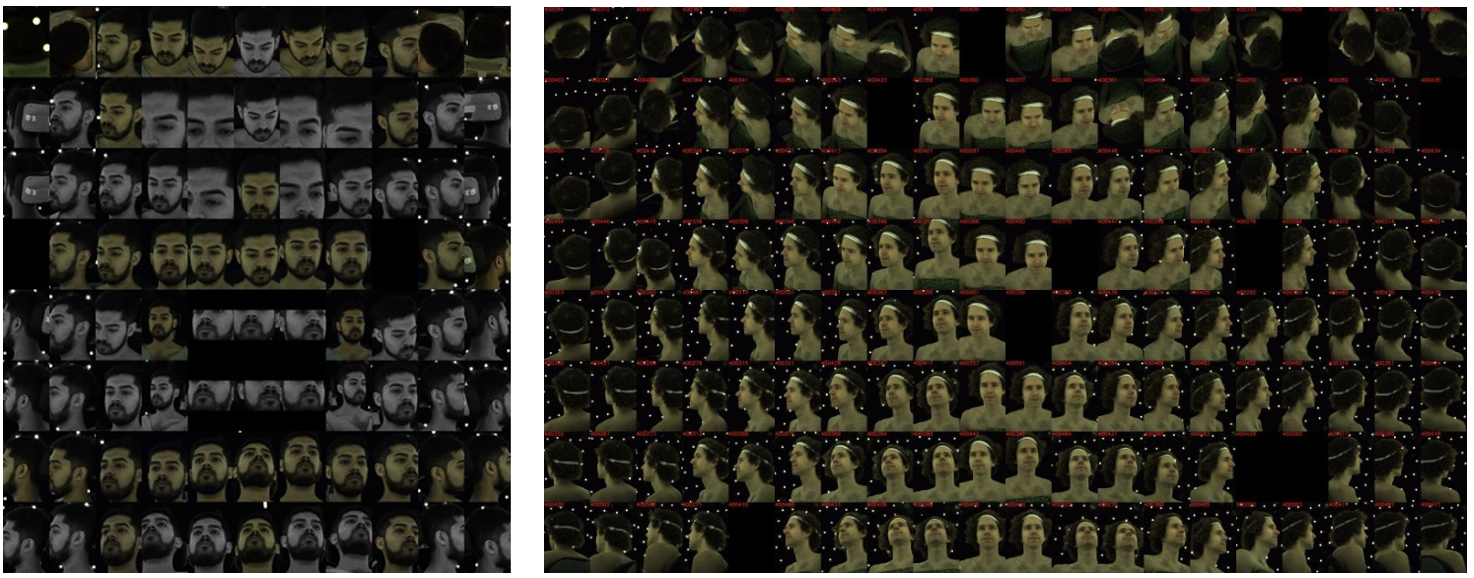}
    \caption{Left: camera views for Mugsy v1. Right: camera views for Mugsy v2.}
    \label{fig:mugsyviews}
\end{figure*}

The cameras are placed on the surface of a sphere with radius 1.2 meters. The cameras all point inward to the middle of the sphere, which is where the head of the participant is located. Figure~\ref{fig:mugsyviews} presents an overview of all the views. The sensors used are IMX253 with pixel size 3.45 $\mu$m. We capture at resolution 4096x2668 (11 megapixels). Shutter speed is at $2.222$ ms. In order for the cameras to capture synchronously, they all listen to the rising edge of a single trigger. While the system is able to capture at 90fps, we only release data at 30fps and downsample the images to 2048x1334 to limit the total dataset size. For lights, we use point light sources that are pointing towards the center of the sphere to illuminate the face of the participant. The lights have diffusers installed to reduce specular highlights on the person’s face and better approximate uniform lighting. All cameras are jointly calibrated using a 3D calibration target~\cite{ha2017deltille} mounted on a robot arm. The calibration process is based on corner detection, intrinsics calibration, and then a final bundle adjustment. Intrinsics and extrinsics of each camera and for each participant are provided as part of the dataset.

% \alex{The cameras are calibrated using a 3D calibration target~\cite{ha2017deltille} mounted on a robot arm. We provide the intrinsics and extrinsics of each camera and for each participant as part of the dataset.}

% \old{
% Once the data is captured, one important step is to jointly  calibrate geometrically the intrinsics and extrinsics of all the cameras. To this end, we leverage a 3D calibration target~\cite{ha2017deltille} mounted on a robot arm to perform calibration. The robot arm will rotate and translate the 3D calibration target in the area where the participant's face originally was while the cameras are capturing. Then, on the captured data from all cameras, we run corner detection, intrinsics calibration, and then a final bundle adjustment to get the intrinsics and extrinsics of all the cameras.}

% \alex{unnecessary information}
% \old{The capture servers are responsible for storing the data generated by the cameras. The cameras can generate up to 1GBps when capturing at 90 fps, hence highly performant servers and disks are needed to be able to save data quickly. We have up to 72 servers in the backend supporting the cameras in Mugsy. Each of the servers have at least 13TB of SSD storage space.}
We release ten captures from the original Mugsy (v1) and three captures obtained with an extended version featuring a significantly larger amount of cameras (Mugsy v2), see Table~\ref{table: mugsy spec} for details. Note that the number of cameras in each capture may be less than the number of cameras shown in the table as individual cameras might fail during a capture.

% \alex{
% We release ten captures from the original Mugsy (v1) and three captures obtained with an extended version featuring a significantly larger amount of cameras (Mugsy v2), see Table~\ref{table: mugsy spec} for details. Note that the number of cameras in each capture may be less than the number of cameras shown in the table as individual cameras might have failed during a capture.
% We downsample the captured images to 2048x1334 for more efficient processing of the captured data.
% }
% \old{For the dataset, it was collected by two generations of Mugsy: V1 and V2. The specs of each generation are shown in ~\ref{table: mugsy spec}. Note that the number of cameras in each capture may be less than the number of cameras shown in the table. This is due to a subset of cameras failing during capture. Furthermore, for Mugsy V1, we only share the images from the color cameras, because the monochrome cameras are not useful for codec avatar training. Also, the dataset only includes images at 30 fps as our processing was done at 30fps. Finally, the released images are of resolution 2048x1334, which is the resolution we use when training our avatars with a differentiable renderer.}

\begin{table*}[t]
  \footnotesize
  \caption{Specifications for each generation of Mugsy.}
  \centering
  \label{table: mugsy spec}
  \begin{tabularx}{\textwidth}{Xrrrrr}
    \toprule
    Mugsy & $\#$ Cameras & Camera resolution & Camera lens & FPS & $\#$ Lights \\
    \midrule
    v1	& 40 color + 50 monochrome & 4096x2668 & 35mm * 2, 85mm * 12, Remaining are 50 mm	& 30/90 &	350\\
    v2	 & 160 color	&	4096x2668 & All 35 mm & 30/90	&	450 \\
    \bottomrule
  \end{tabularx}
\end{table*}

% \begin{figure}[btp] 
% 	\centering
% 	\begin{subfigure}[b]{1.0\linewidth}
%         \includegraphics[width=1\linewidth]{Figures/mugsyv1_multiview.jpg}
%     \end{subfigure}
% 	\caption{A view from different cameras in Mugsy V1.}
% 	\label{fig: mugsy_v1}
% \end{figure}

% \begin{figure}[btp] 
% 	\centering
% 	\begin{subfigure}[b]{1.0\linewidth}
%         \includegraphics[width=1\linewidth]{Figures/mugsyv2_multiview.jpg}
%     \end{subfigure}
% 	\caption{A view from different cameras in Mugsy V2.}
% 	\label{fig: mugsy_v2}
% \end{figure}

\subsection{Capture Script}

\begin{figure}[t] 
	\centering
	\begin{subfigure}[b]{1.0\linewidth}
        \includegraphics[width=1\linewidth]{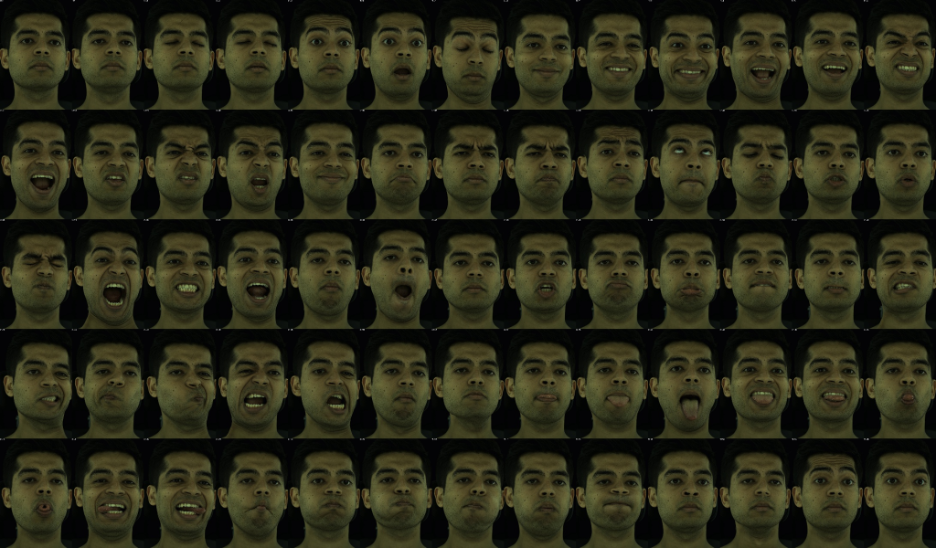}
    \end{subfigure}
	\caption{The 65 peak expressions used in the v1 script. Participants are asked to make their best effort for expressions they may not be able to do, e.g., raise only left or right eyebrow.}
	\label{fig: peack_expression}
\end{figure}

The goal of the face captures is to cover the full range of facial expressions such that a neural avatar like a deep appearance model~\cite{10.1145/3197517.3201401} can learn to interpolate from a range of captured expressions to all possible facial expressions.
To this end, we design a capture script that captures a range of expressions, gaze directions, and 50 phonetically balanced sentences, see Table~\ref{table: mugsy comparision}.
The ten captures from Mugsy v1 follow a script where the expression portion is focused on peak expressions as in Figure~\ref{fig: peack_expression}, which aim to capture motion in different regions of the face independently.
The three captures from Mugsy v2 follow a slightly modified script, where the expression portion is focused on full-face range of motion tasks, and the gaze portion has been simplified.

\begin{table*}[t]
  \footnotesize
  \caption{Comparison between v1 and v2 capture scripts.}
  \centering
  \label{table: mugsy comparision}

  \begin{tabularx}{\textwidth}{rcXcXcl}
    \toprule
    Script & & Expressions & & Gaze & & Sentences \\
    \midrule
    v1	& & 65 peak expressions as shown in Fig.~\ref{fig: peack_expression} \newline Participants go from neutral expression to peak to neutral. Only data from peak to neutral is processed and released.  \newline \newline 1 single range-of-motion segment: ROM07 & & Participants look at 25 fixed markers without turning their head. \newline \newline Participants will look at the leds with normal eyes, wide eyes, squinty eyes, and small head rotations. & & 50 phonetically balanced sentences \\
    \midrule
    v2	& & 2 peak expressions: neutral and eyes closed mouth lightly open. \newline \newline 18 range-of-motion segments covering 118 expressions over the entire face. & & Participants look at 9 cameras without turning their head. When looking at each camera, look at it with normal eyes, wide eyes, squinty eyes, blink, and wink. \newline \newline Additionally, look at a camera and do 5 head rotations. & & 50 phonetically balanced sentences \\
    \bottomrule
  \end{tabularx}

%   \vspace{-4mm}
\end{table*}

\subsection{Data Processing: The Tracking Pipeline}
\label{sec:tracking_pipeline}

\begin{figure*}[t]
  \centering
  \includegraphics[width=\linewidth]{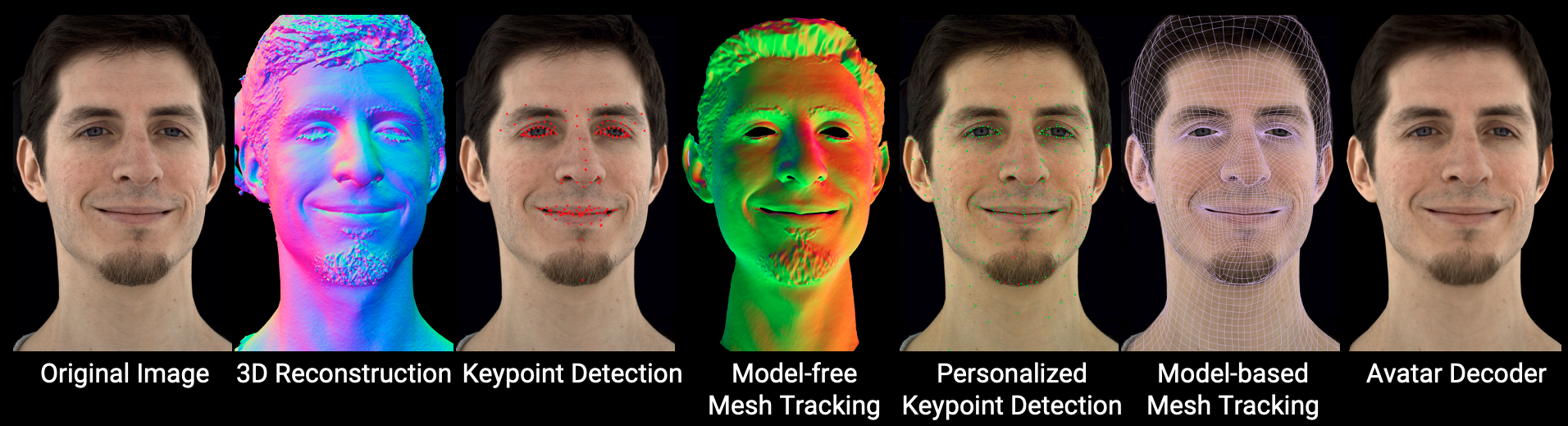}
  \vspace{-7mm}
  \caption{Snapshot of intermediate results from different parts of the pipeline.}
  \label{fig: pipeline}
\end{figure*}

In order to obtain meshes and unwrapped textures from the raw images, we run the captured data through a sophisticated tracking pipeline. Note that we release the resulting meshes and textures, so rebuilding the tracking pipeline is not required for users of this dataset.
The pipeline follows several steps as illustrated in Figure~\ref{fig: pipeline}. First, we run a modified version of parallel Patchmatch~\cite{7410463} on each frame to get a dense 3D mesh reconstruction. Note that dense meshes in different frames do not share the same topology. Next, we detect 2D keypoints on the face. Then, we run sequential tracking with model-free mesh tracking~\cite{10.1145/3414685.3417768} with images, keypoints, and the dense 3D mesh reconstructions as input to get corresponding tracked meshes. Due to the computational cost of these steps, we run this sequential tracking pipeline only on a subset of the captured data, \ie, the expressions and the gaze portion, but not the sentences. 
Given these tracked results, we generate training data for personalized keypoint detection. The personalized keypoints are not just constrained to landmarks where a human annotator can consistently annotate, but also locations on the cheek and forehead, which is hard for a human to annotate consistently, but could be annotated accurately by model-free mesh tracking. The training data is used to train a personalized keypoint detector, which is then used as an initialization of a PCA model-based mesh tracking method. The advantage of using the personalized keypoints and PCA model-based tracking is that sequential tracking is no longer required, and all frames can be tracked in parallel, thus reducing the computational cost significantly and allowing us to process the complete capture efficiently. Finally, once we have the tracked meshes and an image from each view, we unwrap the texture for that specific view to obtain all necessary data for codec avatar generation.

\subsection{Dataset Summary}

In summary, for each of the 13 captured subjects, we provide the following data.

\noindent
\textbf{Raw Images.}
Raw images are directly captured from 40 (v1) to 160 (v2) multi-view cameras at the rate of 30 fps and are released at the resolution of 2048 $\times$ 1334. Raw images can be used as ground truth to compute the screen loss with predicted rendered images from the model.

\noindent
\textbf{Unwrapped Textures.}
Unwrapped textures are provided at a resolution of 1024 $\times$ 1024, and are generated by unwrapping the raw images from the geometry. We wrap each mesh triangle to the corresponding UV texture triangle using barycentric interpolation. Each camera and each frame has its unique view-dependent unwrapped texture.
%\alex{TODO: unwrapped textures at size 1024x1024. View dependent (one unwrapped texture for each camera and frame)}

\noindent
\textbf{Tracked Meshes.}
Meshes are tracked per frame and stored in \textit{.obj} format with same topology. Each mesh consists of 7,306 vertices, with no vertices inside eyes or mouth. By projecting with the provided camera calibrations and headposes, meshes can be aligned with raw images.

\noindent
\textbf{Headposes.}
A headpose is a $3\times 4$ matrix consisting of rotation and translation that represents the rigid body transformation of the head mesh at each frame.

\noindent
\textbf{Audio.}
While most captured expressions are silent, each participant was asked to read 50 phonetically balanced sentences. We provide audio data for these 50 sentences of each participant.

\noindent
\textbf{Metadata.}
The following metadata is provided as well:
\begin{itemize}[leftmargin=*, noitemsep]
    \item \textit{camera calibrations:} we provide each camera's intrinsic and extrinsic matrix.
    \item \textit{frame list:} a list of all frames captured by Mugsy, each line consists of segment name and frame index.
    \item \textit{texture mean:} the mean of the textures across all frames and all cameras.
    \item \textit{texture variance:} the variance of the textures across all frames and all cameras.
    \item \textit{vertex mean:} the mean of the vertices of the meshes across all frames and all cameras.
    \item \textit{vertex variance:} the variance of the vertices of the meshes across all frames and all cameras.
\end{itemize}

\begin{table}[t]
  \footnotesize
  \caption{Mugsy and script version, $\#$ (color) cameras, and $\#$ frames for each released capture.}
  \centering
  \resizebox{\columnwidth}{!}{
  \label{table: capture}
  \begin{tabularx}{0.5\textwidth}{Xrr}
    \toprule
    Capture ID (Mugsy/Script Version) & $\#$ Cameras & $\#$ Frames \\
    \midrule
    m--20171024--0000--002757580--GHS (v1)  & 39 & 9791 \\
    m--20180105--0000--002539136--GHS (v1)  & 39 & 9903 \\
    m--20180226--0000--6674443--GHS	(v1)  & 39 & 12381 \\
    m--20180227--0000--6795937--GHS	(v1)  & 38 & 13478 \\
    m--20180406--0000--8870559--GHS	(v1)  & 40 & 14914 \\
    m--20180418--0000--2183941--GHS	(v1)  & 40 & 15115 \\
    m--20180426--0000--002643814--GHS (v1) & 40 & 10258 \\
    m--20180510--0000--5372021--GHS	(v1)  & 40 & 12619 \\
    m--20180927--0000--7889059--GHS	(v1)  & 34 & 13541 \\
    m--20181017--0000--002914589--GHS (v1) & 34 & 10179 \\
    \midrule
    m--20190529--1004--5067077--GHS	(v2) & 146 & 22928 \\
    m--20190529--1300--002421669--GHS (v2)  & 150 & 24375 \\
    m--20190828--1318--002645310--GHS (v2)  & 147 & 21399 \\
    \bottomrule
  \end{tabularx}
}
\end{table}

%------------------------------------------------------------------------
\section{Technical Details}
\label{sec:technical}

In this section, we demonstrate how to use the dataset and the provided scripts for training. For detailed instructions how to train a codec avatar using our codebase, see the Github repository.

\subsection{Model}
\label{sec:model}

\begin{figure}
    \centering
    \includegraphics[scale=0.1]{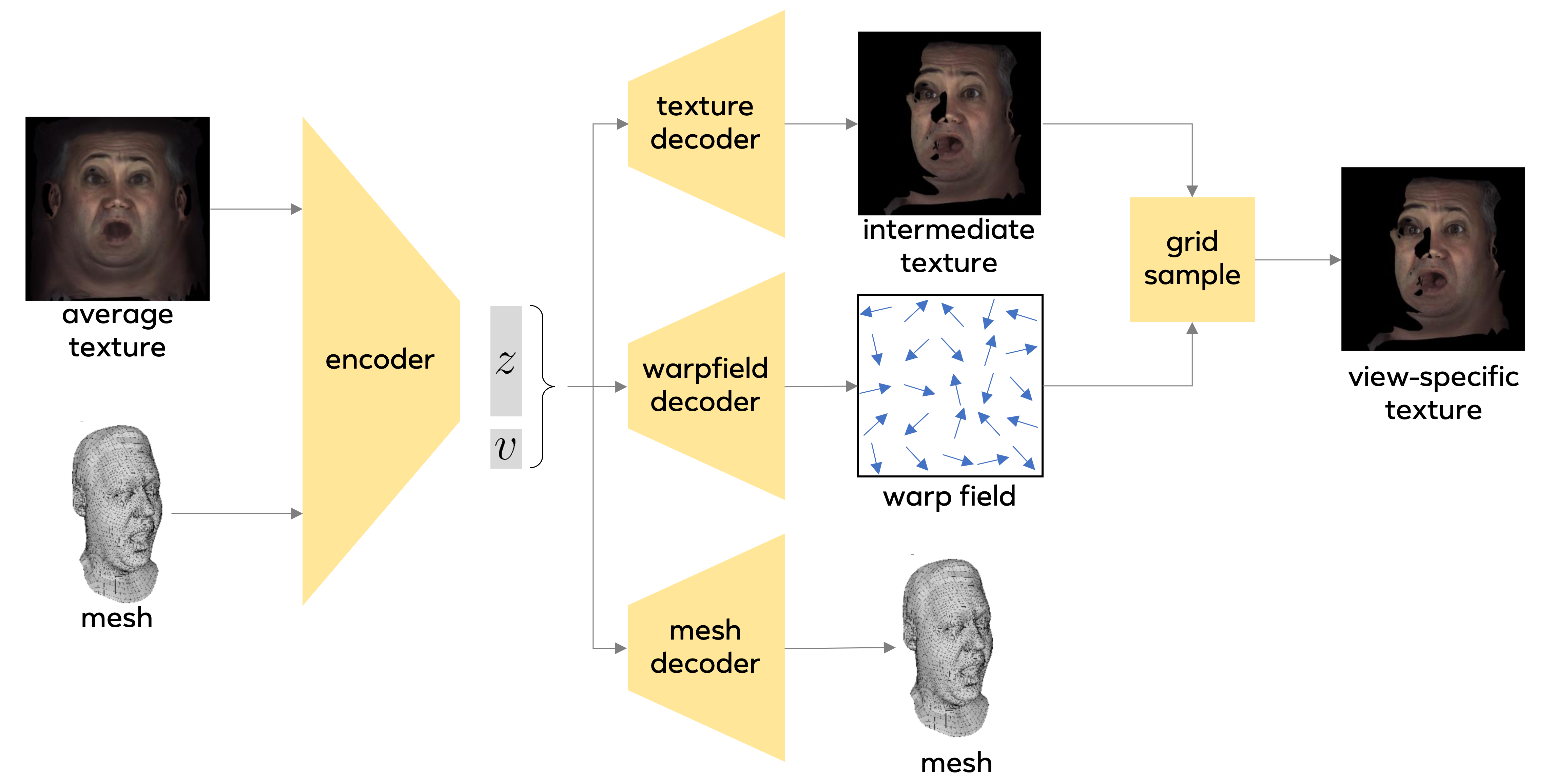}
    \caption{The model consumes, as input, a tracked mesh and an average texture over all camera views and maps these inputs to a latent code, from which a view-dependent texture and the tracked mesh can be decoded.}
    \label{fig:vae}
    \vspace{-5mm}
\end{figure}

Our model greatly resembles the deep appearance model~\cite{10.1145/3197517.3201401}, which, at its core, is a variational autoencoder (VAE) that takes meshes and average texture as input and decodes the view-dependent textures for rendering, see Figure~\ref{fig:vae}. We follow the original algorithm of~\cite{10.1145/3197517.3201401} with some minor exceptions that will be outlined in the following.

As input, the model consumes the average texture from all camera views for a given training frame as well as the tracked mesh.
Using a neural encoder, these inputs are mapped to a 256-dimensional KL-regularized latent space.
Note that the latent space is view-independent as the inputs to the encoder are view-agnostic.
A neural decoder consumes such a view-independent latent representation and a view vector and generates the texture for this specific view as well as the mesh.

The texture encoder consists of 8 convolutional layers, each with kernel size 4 and stride 2, that downsamples the input texture from a resolution of 1024x1024 to a 4x4 feature map. The mesh is encoded with a multi-layer perceptron (MLP) and encoding from texture encoder and mesh encoder are combined into a single 256-dimensional latent vector using a fully connected layer. On the decoder side, the view information is fed into an MLP and the view-feature is concatenated with the latent code. Thus, the texture decoder can be conditioned on this view information and models view-dependent effects in texture space. We explore several different architectures to investigate their generalizing capacity on novel expressions and camera views.

\noindent
\textbf{Color Correction.}
Since different cameras could have different color space, we optimize color correction parameters for each camera. Color correction is performed on the output texture by scaling and adding a bias to each RGB channel. The scaling factors and biases are initialized to 1 and 0, respectively. We fix the color correction parameters of one camera as an anchor and train the other parameters as a part of the model. Applying color correction is necessary, otherwise the reconstruction error will be dominated by the global color difference instead of exact colors of the pixels.

\noindent
\textbf{Spatial Bias.}
For convolutional layers in the decoder for upsampling, instead of adding the same bias value per channel in the feature map, we add a bias tensor that has the same shape as the feature map, meaning that each spatial location has its own bias value. In this way, the model is able to capture more position-specific details in the texture, such as wrinkles and lips.

\noindent
\textbf{Warp Field.}
We can also decode a warp field from the latent space and bilinearly sample the output texture with the warp field. Conceptually, texture generation can be decomposed into two steps: a synthesized texture on a deformation-free template followed by a deformation field that introduces shape variability. Denote $T(p)$ as the value of the synthesized texture at coordinate $p = (x, y)$. Denote $W(p)$ as the estimated deformation field at location $p$. Then, the observed image, $I(p)$, can be reconstructed as follows: $I(p) = T(W(p))$, namely the image appearance at position $p$ is obtained by looking up the synthesized appearance at position $W(p)$. We obtain a warp field in the same way as deformable autoencoders~\cite{dae}: by integrating both vertically and horizontally on the generated warping grid to avoid flipping of relative pixel positions.

\noindent
\textbf{Residual Connection.}
We insert residual layers~\cite{resnet} into our network to make it deeper. We investigate whether this increase in model capacity would make it generalize better.

\begin{figure*}[h!]
  \centering
  \includegraphics[width=\linewidth]{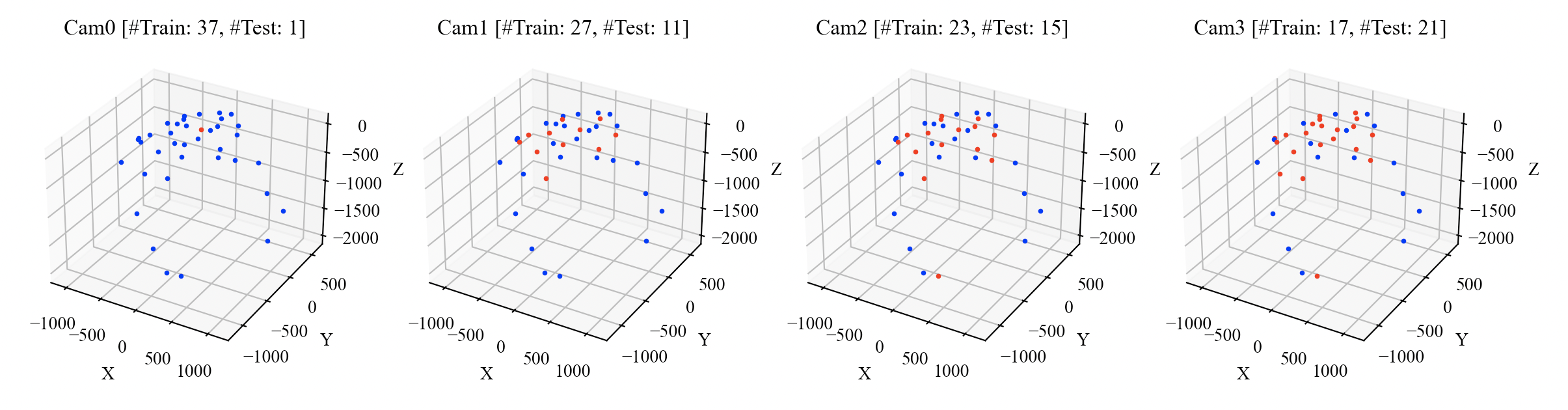}
  \caption{Sets of training and testing camera splits used in ablation study. Red dots represents the position of testing cameras and blue dots represents the position of training cameras. Note that for better visibility of the camera sets, we rendered the plots such that the z-axis is the one pointing away from the identity's face.}
	\label{fig: ablation_camset}
\end{figure*}

\subsection{Training Pipeline}

% \begin{figure}[t]
% 	\centering
% 	\begin{subfigure}[b]{0.6\linewidth}
%         \includegraphics[width=1\linewidth]{Figures/loss_weight_mask.png}
%     \end{subfigure}
% 	\caption{Texture weight mask.}
% 	\label{fig: weight_mask}
% \end{figure}

In~\cite{10.1145/3197517.3201401}, the model was trained with an $\ell_2$-loss on predicted mesh vertices and textures.
We present a set of experiments that diverge from this training strategy by optimizing the screen space loss of the predicted avatars directly against the ground truth images.
To propagate gradients from screen space to the predicted textures and geometries, we use Nvdiffrast~\cite{nvdiffrast} as the differentiable rendering engine.

More formally, given a ground truth image $ I(v) $ captured from a camera at viewpoint $ v $, the loss can be computed by rendering the predicted texture $ \hat T $ and the predicted geometry $ \hat G $ from viewpoint $ v $ using the differentiable rendering function $ R $, and comparing it to the respective ground truth image $ I(v) $,
\begin{align}
    L = \|R(v, \hat T, \hat G) - I(v) \|^2.
\end{align}
However, naively computing an $\ell_2$-loss from the 3D rendered avatar in screen space and the ground truth images is not reasonable:
ground truth images include background that must not be learned by the model.
Moreover, humans are more sensitive to changes in around the eyes and mouth. To mitigate these two issues, we apply a \textit{foreground mask $ F $} (defined in texture space) calculated from non-background pixels in image space and assign a higher weight to eyes and mouth regions during training than to the remaining face regions using a manually created \textit{texture weight mask} $ M $. % (Figure~\ref{fig: weight_mask}.
We therefore render both the predicted geometry $ \hat G $ and the predicted texture $ \hat T $ into screen space, but also the weight mask. The resulting loss then is defined as
\begin{align}
    L = \|R(v, M, \hat G) \odot R(v, \hat T, \hat G) \odot F - I(v) \odot F \|^2,
\end{align}
where $ \odot $ denotes element-wise multiplication.

We add an explicit loss between predicted and (tracked) ground truth geometry to enhance learning the right geometric shape of the face during training as well as the KL loss for the VAE.
The combined loss term is
\begin{align}
    L = & \lambda_1 \cdot \|R(v, M, \hat G) \odot R(v, \hat T, \hat G)\odot F - I(v) \odot F \|^2 \nonumber \\
        + & \lambda_2 \cdot \|\hat G - G\|^2 \nonumber \\
        + & \lambda_3 \cdot \mathrm{KL}(\mathcal{N}(\mu_z, \sigma_z) || \mathcal{N}(0, I)),
\end{align}
with $ \mu_z $ and $\sigma_z $ being the predicted mean and variance of the latent distribution. Here the images and textures are normalized by per-pixel mean and variance during training. For faster convergence and unequal learning rate, we multiply the output mean of the encoder by $0.1$ and the log standard deviation by $0.01$. We use Adam~\cite{adam} as optimizer and perform 200K iterations for all the experiments. We set $\lambda_1 = \lambda_2 = 1$ and $\lambda_3=0.01$.

%------------------------------------------------------------------------

\section{Experiments}\label{sec:results}

In the following, we evaluate the changes to the original model from~\cite{10.1145/3197517.3201401} that have been suggested in Section~\ref{sec:model}.

\subsection{Experiment Setting}

The ablation study is performed on a single identity, m--20180227--0000--6795937--GHS, and with varying training and test camera views.
To evaluate the effect of the number of camera views available during training, we run each experiment with four different settings: (1) training on 37 cameras, (2) training on 27 cameras, (3) training on 23 cameras, and (4) training on 17 cameras.
Figure~\ref{fig: ablation_camset} illustrates the location of the train cameras for these four settings. In total, training takes around one day for each avatar using a P3.16x instance with eight Nvidia V100 GPUs on AWS.

% \paragraph{Camera Split}
% We perform ablation study on the captured data on one identity $\texttt{m--20180227--0000--6795937--GHS}$ to test different architectures on novel views and expressions. We manually split the testing and training camera sets from the total of 38 cameras such that the testing cameras are approximately equally spaced. There are 4 sets of camera splits: cam0, cam1, cam2, and cam3, we used for the ablation study, where cam0 has the greatest number of training cameras (37) and least number of testing cameras (1), and Cam3 has the least number of training cameras (17) and greatest number of testing cameras (21). Figure \ref{fig: ablation_camset} shows the position of training and testing cameras in each split and table \ref{table: ablation_camset} in Appendix \ref{sec:appendix_d} lists the exact training and testing cameras in each split. We evaluate the models on all frames in $\texttt{ROM}$ expressions.

\begin{figure*}[h!]
     \centering
     \begin{subfigure}[t]{0.32\textwidth}
         \centering
         \includegraphics[width=\textwidth]{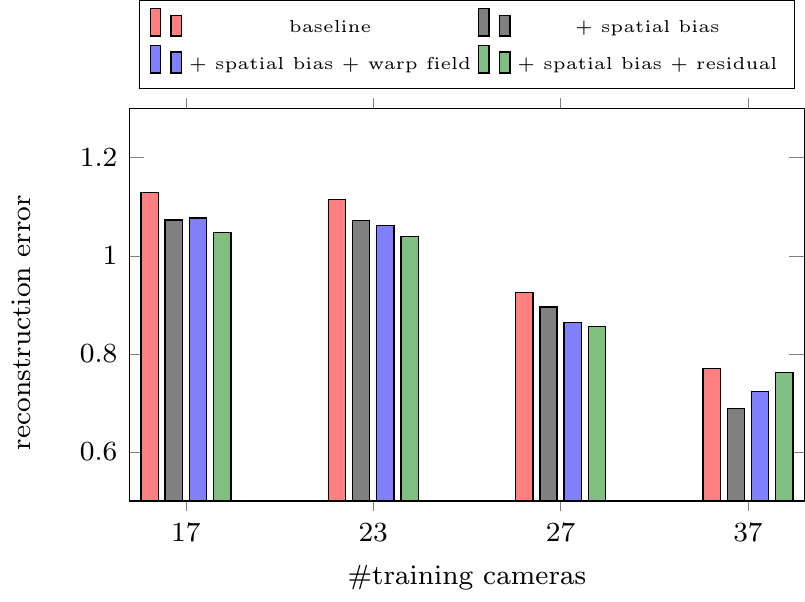}
         \caption{novel view synthesis}
         \label{fig:novelview}
     \end{subfigure}
     \hfill
     \begin{subfigure}[t]{0.32\textwidth}
         \centering
         \includegraphics[width=\textwidth]{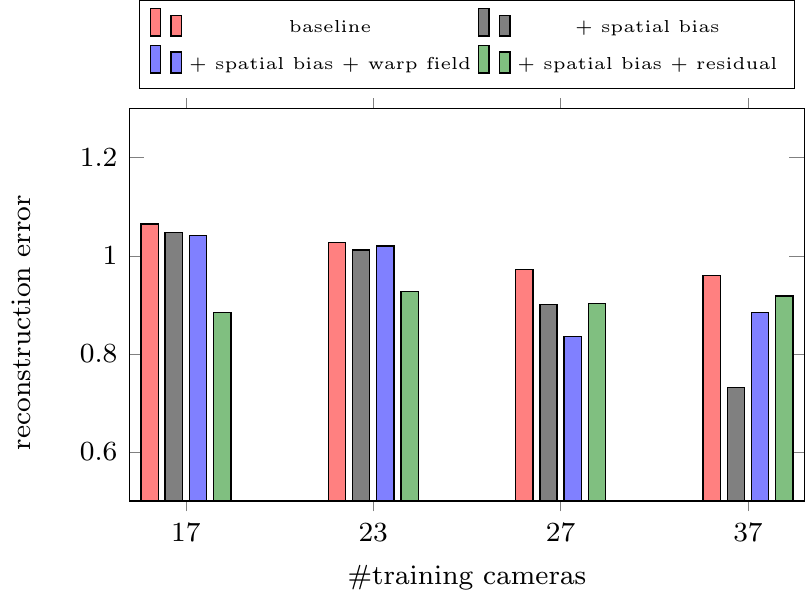}
         \caption{novel expression synthesis}
         \label{fig:novelexpression}
     \end{subfigure}
     \hfill
     \begin{subfigure}[t]{0.32\textwidth}
     \centering
         \includegraphics[width=\textwidth]{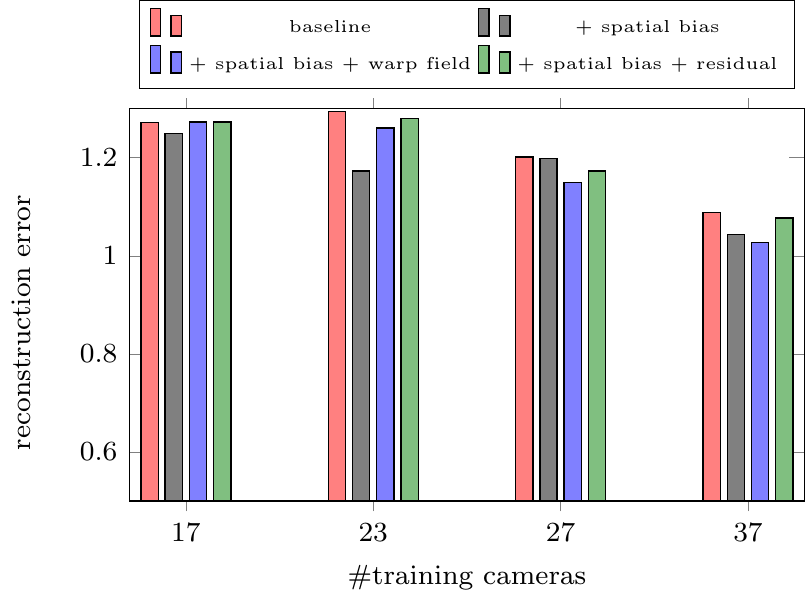}
         \caption{novel view + novel expression}
         \label{fig:joint}
     \end{subfigure}
     \vspace{0.3cm}
     
    \centering
    \footnotesize
    \begin{tabularx}{\textwidth}{Xrrrr|rrrr|rrrr} % transformed original values with 10.686 ** 2 * x / 100.0 to be back in pixel intensity space
        \toprule
                                    & \multicolumn{4}{c}{(a) novel view synthesis}        &  \multicolumn{4}{c}{(b) novel expression synthesis}        & \multicolumn{4}{c}{(c) novel view+expression synthesis} \\
                                      \cmidrule(lr){2-5} \cmidrule(lr){6-9} \cmidrule(lr){10-13}
        \# training cameras         & 17     & 23     & 27     & 37            & 17      & 23        & 27        & 37            & 17        & 23        & 27        & 37          \\
        \midrule
        baseline                    & $ 1.129 $ & $ 1.115 $ & $ 0.925 $ & $ 0.770 $ & $ 1.065 $ & $ 1.027 $ & $ 0.972 $ & $ 0.960 $ & $ 1.272 $ & $ 1.294 $ & $ 1.202 $ & $ 1.088 $ \\
        + spatial bias              & $ 1.073 $ & $ 1.072 $ & $ 0.896 $ & $ 0.689 $ & $ 1.048 $ & $ 1.012 $ & $ 0.901 $ & $ 0.732 $ & $ 1.250 $ & $ 1.173 $ & $ 1.198 $ & $ 1.044 $ \\
        + spatial bias + warp-field & $ 1.077 $ & $ 1.062 $ & $ 0.864 $ & $ 0.724 $ & $ 1.041 $ & $ 1.020 $ & $ 0.835 $ & $ 0.885 $ & $ 1.273 $ & $ 1.261 $ & $ 1.150 $ & $ 1.027 $ \\
        + spatial bias + residual   & $ 1.048 $ & $ 1.040 $ & $ 0.856 $ & $ 0.762 $ & $ 0.885 $ & $ 0.927 $ & $ 0.903 $ & $ 0.918 $ & $ 1.273 $ & $ 1.280 $ & $ 1.173 $ & $ 1.077 $ \\
        \bottomrule
    \end{tabularx}
    \caption{Reconstruction errors for four architectures trained on 17, 23, 27, and 37 camera views. We compare system performance for novel view synthesis, novel expression synthesis, and joint synthesis of novel expressions in novel views. The reconstruction errors refer to pixel intensity difference.}
    \vspace{-4mm}
\end{figure*}
We conduct ablation studies for four different model architectures: (1) the model from~\cite{10.1145/3197517.3201401} without the spatial bias, (2) the original model from~\cite{10.1145/3197517.3201401} including spatial bias, (3) the model plus spatial bias and warp field, and (4) the model with spatial bias and residual connections, as described in Section~\ref{sec:model}.

Building photorealistic 3D facial avatars requires models to accurately produce novel views on the avatar (novel view synthesis), and to generalize well to facial expressions that are unseen in the training set (novel expression synthesis).
For novel view synthesis, we evaluate the reconstruction error on held out cameras (red dots in Figure~\ref{fig: ablation_camset}) and expressions seen during training, and for novel expression synthesis we evaluate the reconstruction error on camera views used during training (blue dots in Figure~\ref{fig: ablation_camset}) but with held out facial expressions.
We also examine both properties jointly by evaluating on held out cameras and held out expressions.

% To examine the model's generalization ability, we have 3 sets of evaluations which will tell us the model's interpolation capacity on either novel viewpoints, novel expressions, or on both novel viewpoints and expressions: 
% \begin{enumerate}
% \item Testing Camera on Testing Expression (Generalization on Viewpoint + Expression)
% \item  Testing Camera on Training Expression (Generalization on Viewpoint)
% \item  Training Camera on Testing Expression (Generalization on Expression)
% \end{enumerate}
For fair comparison, we train color correction parameters for unseen testing cameras for two epochs on the validation data before evaluation. We also fine-tune the encoder on the testing expression as we would like to test the capacity of the conditional decoder independent of the encoder. To enable the encoder to generate the latent code that is specialized to the given dataset, we therefore train the encoder on the validation data while freezing the decoder weights on given cameras for 10 epochs before evaluation.

\subsection{Results and Analysis}

\noindent
\textbf{Novel View Synthesis.}
We first evaluate the model with respect to its ability to synthesize novel views.
In Figure~\ref{fig:novelview}, we plot the screen error measured on held out cameras over the number of available training cameras.
The expressions we evaluate on are seen during training but the testing cameras are at unseen positions.
The higher the number of available training cameras, the lower the reconstruction error.
Spatial biases in the network are crucial for high accuracy in novel view synthesis as they encode view-independent texture information, compare the baseline without spatial bias (red bar) to all other architectures that have spatial biases.
An increased number of layers achieved through the use of residual connections is particularly important for a lower number of training cameras (green bar).

\noindent
\textbf{Novel Expression Synthesis.}
To evaluate the model's ability to generate unseen facial expressions, we therefore evaluate on seen views (\ie, training cameras) only, see Figure~\ref{fig:novelexpression}. As before, a lack of spatial biases leads to a significant degradation of reconstruction quality compared to other architectures that use spatial biases. While the baseline model plus spatial bias performs best when dense training views are available, it quickly deteriorates for fewer available training cameras.
The deeper model with residual connections shows the most consistent performance, being at a rather low reconstruction error independent of the number of available training cameras.

\noindent
\textbf{Joint View- and Expression Synthesis.}
Last, we combine the two previous settings and evaluate the performance of our four architectures on unseen facial expressions and novel camera views simultaneously, see Figure~\ref{fig:joint}. As before, the larger the number of available training camera views, the smaller the reconstruction error. Note, however, that each individual task alone is easier to solve than the joint task: the reconstruction errors for novel view synthesis and novel expression synthesis alone are significantly lower than the reconstruction errors for the joint task of novel view- and expression synthesis.

\begin{figure}[ht!]
  \centering
  \includegraphics[width=\linewidth]{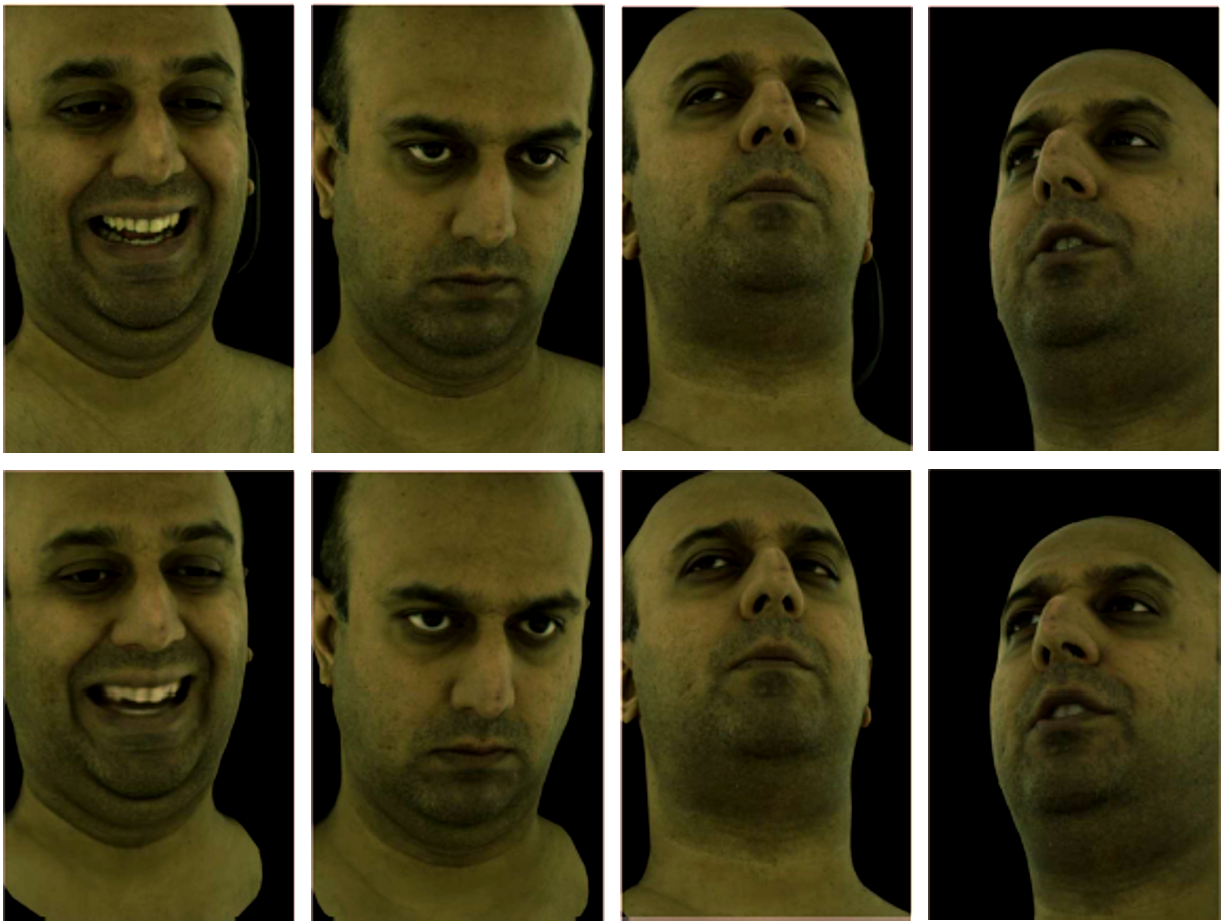}
  \caption{Reconstruction result of the warp-field model using 37 camera views during training. Top row is ground-truth, bottom row is the 3D reconstruction generated by the model.}
	\label{fig: recons}
 \vspace{-5mm}
\end{figure}

\noindent
\textbf{Qualitative Results.}
Figure \ref{fig: recons} shows predicted frames after training of a warp-field model. The top row shows ground-truth and the bottom shows rendered image using predicted mesh and texture. Although the reconstruction generally looks good, the model struggles to predict high frequency details such as teeth and eye-lashes.

\section{Conclusion}
We release a large-scale multi-view codec-avatar dataset for neural face rendering, along with training, evaluation, and visualization code and pretrained models for 13 identities.
Besides the dataset, to understand how different model architectures respond to interpolating on unseen viewpoint and expression, we conduct an ablation study and identify that the baseline VAE model benefits from adding spatial bias, texture warp field and residual connections. We hope that this dataset will push the limit of facial reconstruction further and facilitate future research of VR telepresence.

\newpage
%%%%%%%%% REFERENCES
{\small
\bibliographystyle{ieee_fullname}
\bibliography{main}
}

\end{document}